# Applications of Machine Learning Techniques in Human Activity Recognition

Jitenkumar B Rana
Tanya Jha
Rashmi Shetty

**Abstract**

Human activity detection has seen a tremendous growth in the last decade playing a major role in the field of pervasive computing. This emerging popularity can be attributed to its myriad of real-life applications primarily dealing with human-centric problems like healthcare and eldercare. Many research attempts with data mining and machine learning techniques have been undergoing to accurately detect human activities for e-health systems. This paper reviews some of the predictive data mining algorithms and compares the accuracy and performances of these models. A discussion on the future research directions is subsequently offered.

## I. Introduction

Advancements in the field of medicine have greatly improved our quality of life which is clearly evident in the rise of life expectancy. Rising health-care costs, especially in the treatments of the elderly, have called for cost-cutting measures from various health-care institutes. Technological advancements could contribute significantly in cutting down health-care costs by making the medical staff and the hospital environment more efficient.

In the last decade, Human Activity Recognition (HAR) has emerged as a powerful technology with the potential to benefit elderly and differently-abled. Simple human activities have been successfully recognized and researched so far. Recognizing complex human activities still remain challenging and active research is being carried out in this area. The primary goal of HAR is to accurately detect common human activities in real-life settings. Many data mining techniques like Decision Tree, Random Forest, AdaBoost and Support Vector Machine have been used to predict activities with good accuracy. In this paper, some of these HAR classifier models are applied on a dataset which is available free on [1]. A comparison of the performance and accuracy of the models is provided.

This paper is organized as follows:  Section 2 describes the data collection technique and elaborates on the building of the classifier. Section 3 is the review of the commonly used data mining and machine learning techniques and a comparison of the performance of different models. Section 4 concludes the paper with future research directions.

## II. Building a Classifier for Wearable Accelerometers' Data

To build a classifier for the data from the 4 accelerometers steps followed were: data collection, feature extraction, feature selection. Initial data set was randomly sampled into training set (70%) and testing set (30%) to estimate out of sample accuracy of the model.

**2.1 Data Collection:**

For our research exercise, we have used the dataset available freely on [1]. The dataset is a collection of accelerometer readings from 4 sensors (on belt, on left thigh, on right ankle, on right arm) worn by each of 4 healthy subjects while performing certain activities in 5 different ways (sitting down, standing up, walking, standing, sitting) for in total of 8 hours. Each activity was performed separately by the subjects. The outcome class to predict was the way in which activity was performed (sitting, standing, standing up, sitting down, and walking).

**2.2 Feature Selection:**

Following features were selected for model building. The list of best features to be used for model was obtained from [2]. The final features used for model are as follows:

(1) Sensor on the Belt: discretization of the module of acceleration vector, variance of pitch, and variance of roll;
(2) Sensor on the left thigh: module of acceleration vector, discretization, and variance of pitch;
(3) Sensor on the right ankle: variance of pitch, and variance of roll;
(4) Sensor on the right arm: discretization of the module of acceleration vector; From all sensors: average acceleration and standard deviation of acceleration.

Variance of different pitch and roll angles and average acceleration and standard deviation were calculated using moving average with window length of 9. Accelerometer reading can be considered as time series data. The length 9 on moving window was arrived at observing the spikes in Auto-Correlation and Partial Autocorrelation of the data.

## III. Machine Learning Techniques and their application in HAR

In past few decades, there has been drastic change in the way, data is stored, perceived and processed. Tremendous amount of data is generated every second and if this data is used and analyzed efficiently, it can reveal very important insights. Lot of data mining techniques has evolved in analyzing the huge amount of data.

One important part of the prediction is the selection of suitable models. In our exercise, we have built models using several machine learning techniques and compared the accuracy of different algorithms. The machine learning algorithms used are as follows. We used R-programming tool (version 3.2.2) for computation and implementation of the model.

**(1). Decision Trees**

Decision trees are one of the common algorithms for classification problems such as Human Activity Recognition. First model was built using C4.5 decision tree for classification. Decision Trees are easy to understand. However, if there is a non-linear relationship between predictors and outcome, accuracy will suffer.

**(2). Adaptive Boost (AdaBoost) for Multiclass**

AdaBoost is a performance boosting technique. This algorithm tends to assign more importance to the incorrectly classified examples by weak learners and so challenges weak learners to perform well. We used AdaBoost technique with 10 decision trees to improve classification accuracy of a single deep classification tree.

**(3). Random Forest**

Random forest is also the algorithm which tends to combine weak learners to improve accuracy. It bootstraps different predictors and builds multiple weak trees from bootstrapped predictors. Bootstrapping of predictors ensures less correlated trees. And finally it combines weak decision trees to predict the outcome. This algorithm also yields much better classification accuracy over decision trees.

**(4). Support Vector Machine**

It is also a supervised learning algorithm. SVM model represents samples as points in space and separates the points based on outcome categories with a clear dividing gap. The category of new points is determined based on which side of the gap they fall on.

Here is the comparison of the performances of 4 algorithms used above:

| Sr | Algorithm Used | No of Trees | Classification Accuracy in % |
|---|---|---|---|
| 1 | Decision Tree | 1 | 86.86 |
| 2 | Boosted Decision Tree (AdaBoost) | 10 | 99.40 |
| 3 | Random Forest | 10 | 99.80 |
| 4 | Support Vector Machine | NA | 98.21 |

## IV. Conclusion

We compared four algorithms for the implementation. We could see that using advanced ensembling techniques or techniques such as Support Vector Machine improves accuracy by a large extent over single decision tree. We see Random Forest gives the most accurate result with an accuracy of 99.8 followed by AdaBoost with 99.4.